\documentclass[11pt,a4paper]{article}
\usepackage[hyperref]{emnlp2020}
\usepackage{times}
\usepackage{latexsym}

\usepackage{microtype}
\usepackage{hyperref}
\usepackage{url}
\definecolor{ao(english)}{rgb}{0.0, 0.5, 0.0}
\newcommand{\ignore}[1]{}

\newcommand{\cbr}{\textsc{CBR}\xspace}
\newcommand{\fb}{FB122\xspace}
\newcommand{\nell}{NELL-995\xspace}
\newcommand{\wn}{WN18RR\xspace}
\newcommand{\grinch}{\textsc{Grinch}\xspace}

\usepackage{amsthm}

\usepackage[multiple]{footmisc}

\usepackage{pslatex}
\usepackage{latexsym}
\usepackage{amsfonts}
\usepackage{mathtools}
\usepackage{graphicx}
\usepackage{amsmath}
\usepackage{bbm}
\usepackage{xspace}
\usepackage{tikz-dependency}
\usepackage{balance}
\usepackage{enumitem}
\usepackage{mathtools}
\usepackage{algorithm}
\usepackage{algorithmicx}
\usepackage{eqparbox}

\usepackage{algcompatible}

\usepackage{footnote}
\makesavenoteenv{tabular}

\usepackage{varwidth}
\usepackage{pifont}
\usepackage{float}
\usepackage{framed} % for boxes
\usepackage{verbatim}
\usepackage{pgfplots}
\usepackage[rightcaption]{sidecap}
\usepackage{subfig}
\usepackage{wrapfig}
\usepackage{multirow}
\DeclareSymbolFont{extraup}{U}{zavm}{m}{n}
\DeclareMathSymbol{\varheartsuit}{\mathalpha}{extraup}{86}

\usepackage{xargs} % commandx 
\usepackage[colorinlistoftodos,prependcaption,textsize=tiny]{todonotes}

\usepackage{marginnote}
\usepackage{lipsum}

\usepackage{microtype}
\usepackage{booktabs}
\usepackage{rotating}
\aclfinalcopy % Uncomment this line for the final submission

\newcommand{\ent}{\ensuremath{e}}
\newcommand{\rel}{\ensuremath{r}}
\newcommand{\pth}{\ensuremath{p}}
\newcommand{\stpath}{\ensuremath{\textsf{st}(\pth)}}
\newcommand{\enpath}{\ensuremath{\textsf{en}(\pth)}}
\newcommand{\lenpath}{\ensuremath{\textsf{len}(\pth)}}
\newcommand{\typepath}{\ensuremath{\textsf{type}(\pth)}}

    \title{Probabilistic Case-based Reasoning for \\Open-World Knowledge Graph Completion}

\author{
Rajarshi Das, Ameya Godbole, Nicholas Monath, Manzil Zaheer, Andrew McCallum\\
University of Massachusetts, Amherst, USA\\
Google Research, USA\\
\texttt{\{rajarshi, agodbole, nmonath, mccallum\}@cs.umass.edu}\\
\texttt{manzilzaheer@google.com}
}

\date{}

\begin{document}
\maketitle
\begin{abstract}

A case-based reasoning (CBR) system solves a new problem by retrieving `cases' that are similar to the given problem. 
If such a system can achieve high accuracy, it is appealing owing to its simplicity, interpretability, and scalability.
In this paper, we demonstrate that such a system is achievable for reasoning in knowledge-bases (KBs).
Our approach predicts attributes for an entity by gathering reasoning paths from similar entities in the KB.
Our probabilistic model estimates the likelihood that a path is effective at answering a query about the given entity. 
The parameters of our model can be efficiently computed using simple path statistics and require no iterative optimization.
Our model is non-parametric, growing  dynamically as new entities and relations are added to the KB. 
On several benchmark datasets our approach significantly outperforms other rule learning approaches and performs comparably to state-of-the-art embedding-based approaches. 
Furthermore, we demonstrate the effectiveness of our model in an ``open-world'' setting where new entities arrive in an online fashion, significantly outperforming state-of-the-art approaches and nearly matching the best offline method.\footnote{Code available at \url{https://github.com/ameyagodbole/Prob-CBR}}

\end{abstract}
\section{Introduction}
\label{sec:intro}
We live in an evolving world with a lot of heterogeneity as well as new entities being created continuously.
For example, scientific papers and Wikipedia pages describing facts about new entities, are being constantly added (e.g. \textsc{Covid-19}). 
These new findings further trigger the inference of newer facts, each with its own diverse reasoning. We are interested in developing such automated reasoning systems for large knowledge-bases (KBs).

In machine learning, non-parametric methods hold the promise of handling evolving data \cite{cover1967nearest,rasmussen2000infinite}. Most current KG completion models learn low dimensional parametric representation of entities and relations via tensor factorization or sophisticated neural approaches \cite{nickel2011three,bordes2013translating,socher2013reasoning,sun2019rotate,vashishth2020compositionbased}. Another line of work learns Horn-clause style reasoning rules from the KG and stores them in its parameters \cite{ntp,das2018go,minervini2019differentiable}. However, these parametric approaches work with a fixed set of entities and it is unclear how these models will adapt to new entities.
\begin{figure*}
    \centering
    \includegraphics[width=2\columnwidth]{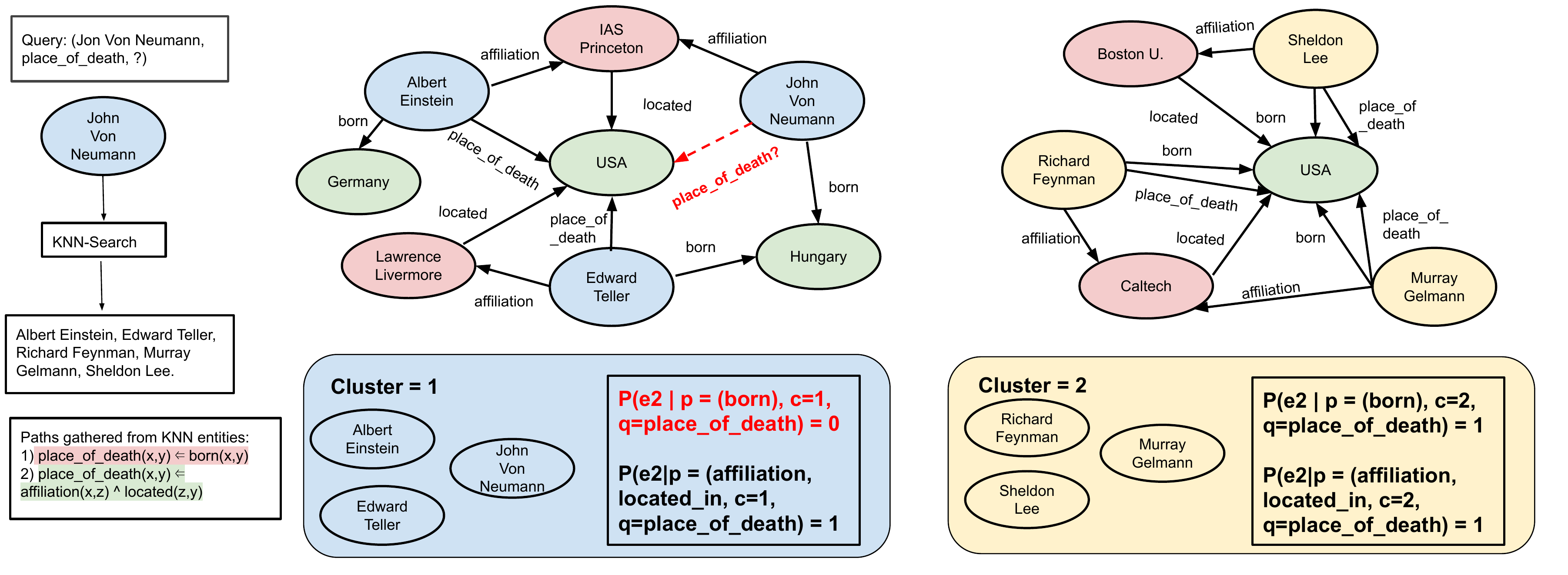}
    \caption{Given the query, (\textsc{Jon Von Neumann},~\textsc{place\_of\_death},~?), our model gathers reasoning paths from similar entities such as other scientists. However, not all gathered paths work for a query e.g. the path (`\textsc{born}(x, y)') would not work for \textsc{Von Neumann}. This highlights the importance of learning path weights for \emph{clusters of similar entities}. Even though `\textsc{born\_in}' could be a reasonable path for predicting \textsc{place\_of\_death}, this does not apply for \textsc{Von Neumann} and other scientists in his cluster. The precision parameter of the path given the cluster helps in penalizing the \textsc{`born\_in'} path. Note that the node \textsc{USA} is repeated twice in the figure to reduce clutter.}
    \label{fig:intro}
\end{figure*}

This paper presents a $k$-nearest neighbor (KNN) based approach for KG reasoning that is reminiscent of case-based reasoning (CBR) in classical AI. A CBR system solves a new problem by retrieving `cases' that are similar to the given problem, revising the solution to retrieved cases (if necessary) and reusing it for the new problem \citep[inter-alia]{schank1982dynamic,leake1996cbr}. For the task of finding a target entity given a source entity and binary KG relation (e.g. (\textsc{John Von Neuman},~\textsc{place\_of\_death},~?) in Figure~\ref{fig:intro}), our approach first retrieves $k$ similar entities (cases) to the query entity. Next, for each retrieved entity, it finds multiple KG paths\footnote{A path is a contiguous sequence of KG facts such as  \textsc{Richard Feynman} $\to$ \textsc{affiliated} $\to$ \textsc{Caltech} $\to$ \textsc{located} $\to$ \textsc{USA}.} (each path is a solution to retrieved cases) to the entity they are connected by the query relation (e.g. paths between (\textsc{Richard Feynman, USA})). However, one solution seldom works for all queries. For example, even though the path `\textsc{born\_in}' is predictive of `\textsc{place\_of\_death}' for US-born scientists (figure~\ref{fig:intro}), it does not work for scientists who have immigrated to USA. To handle this, we present a probabilistic CBR approach which learns to weighs paths with respect to an estimate of its prior and its precision, given the query. The prior of a path represents its frequency while the precision represents the likelihood that the path will lead to a correct answer entity. To obtain robust estimates of the path parameters, we cluster similar entities together and compute them by simple count statistics (\S\ref{sub:param_estimation}).

Apart from computing these estimates, our method needs \emph{no further training}. Overall, our simple approach outperforms several recent parametric rule learning methods \cite{das2018go,minervini2019differentiable} and performs competitively with various state-of-the-art KG completion approaches \cite{dettmers2018convolutional} on multiple datasets. 

An advantage of non-parametric models is that it can adapt to growing data by adjusting its number of parameters. In the same spirit, we show that our model can seamlessly handle an `open-world' setting in which \emph{new entities} arrive in the KG. This is made possible by several design choices such as (a) representing entities as sparse (non-learned) vector of its relation types (\S\ref{sub:contextual_entities}), (b) our use of an online non-parametric hierarchical clustering algorithm  \cite{monath2019scalable} that can efficiently recompute changes in cluster assignments because of the newly added entity (\S\ref{sub:open_world}), (c) and a simple and efficient way of recomputing the prior and precision parameters for paths per cluster (\S\ref{sub:param_estimation}).

Current models for KG completion that learn entity representations for a fixed set of entities cannot handle the open-world setting. In fact we show that, retraining the models continually with new data leads to severe degradation of the model performance with models forgetting what it had learned before.  For example, the performance (MRR) of \textsc{Rotate} model \cite{sun2019rotate} drops by 11 points (absolute) on \wn in this setting (\S\ref{sub:online}). On the other hand, we show that with new data, the performance of our model is consistent as it is able to seamlessly reason with the newly arrived data.

Our work is most closely related to a recent concurrent work by \citet{cbr} where they propose a model that gathers paths from entities similar to the query entity. However, \citet{cbr} encourages path that occur frequently in the KG and does not learn to weigh paths differently for queries.  This often leads to wrong inference leading to low performance. For example,  on the test-II evaluation subset of \fb where all triples can be inferred by logical rules,  \citet{cbr} scores quite low (63 MRR) because of learning incorrect rules. On the other hand, we score significantly higher (94.83 MRR) demonstrating that we can learn more effective rules. In fact, we consistently and significantly outperform \citet{cbr} on several benchmark datasets. Also, unlike us, they do not test themselves in the challenging open-world setting. 
% and it is unclear how much they will generalize in that setting.

The contributions of this paper are as follows: (a) We present a KNN based approach for KG completion that gathers reasoning paths from entities that are similar to the query entity. Following a principled probabilistic approach (\S\ref{sub:non_param_reasoning}), our model weighs each path by its likelihood of reaching a correct answer which penalizes paths that are spurious in nature. (b) The parameters of our model grow with data and can be estimated efficiently using simple count statistics (\S\ref{sub:open_world}). Apart from this, our approach needs \emph{no training}. We show that our simple approach significantly outperforms various rule learning methods \cite{das2018go,minervini2019differentiable,cbr} on many benchmark datasets. (c) We also show that our model can easily handle addition of facts about new entities and is able to seamlessly integrate and reason with the newly added data significantly outperforming parametric embedding based models.

\section{Non-parametric Reasoning in KGs}
\label{sec:model}
\subsection{Notation and Task Description}
\label{sub:notation}
 Let $\mathcal{V}$ denote the set of entities, $\mathcal{R}$ denote the set of binary relations and $\mathcal{G}$ denote a KB or equivalently a Knowledge Graph (KG). Formally, $\mathcal{G} = (\mathcal{V}, E, \mathcal{R})$ is a directed labeled multigraph where $\mathcal{V}$ and $E$ denote the vertices and edges of the graph respectively. Note that, $E \subseteq \mathcal{V} \times \mathcal{R} \times \mathcal{V}$. Let $\left(\ent_{1}, \rel, \ent_{2}\right)$ denote a fact in $\mathcal{G}$ where $\ent_{1}$, $\ent_{2}$ $\in$ $V$ and $\rel \in E$. Also, following previous approaches \citep{bordes2013translating}, we add the inverse relation of every edge, i.e., for an fact $(\ent_{1}, \rel, \ent_{2})\in E$, we add the edge $(\ent_{2}, \rel^{-1}, \ent_{1})$ to the graph. (If the set of binary relations $\mathcal{R}$ does not contain the inverse relation $\rel^{-1}$, it is added to $\mathcal{R}$ as well).

\textbf{Task:} We consider the task of query answering on KGs,  i.e., answering questions of the form $\left(\ent_{1q}, \rel_{q}, ?\right)$, where answer is an entity in the KG.

\textbf{Paths in KG:} A path in a KG between two entities $\ent_{s}$, $\ent_{t}$ is defined as a sequence of alternating entity and relations that connect $\ent_{s}$ and $\ent_{t}$. A length of a path is the number of relation (edges) in the path. Formally, let a path $\pth = (\ent_{1}, \rel_1, \ent_{2}, \ldots, \rel_{n}, \ent_{n+1})$ with $\stpath = \ent_{1}$, $\enpath = \ent_{n+1}$ and $\lenpath = n$. We also define a \emph{path type} as the sequence of the relations in $p$, i.e., $\typepath = (\rel_1, \rel_2, \ldots, \rel_n)$. Let $\mathcal{P}$ denote the set of all paths in $\mathcal{G}$. Let $\mathcal{P}_n \subseteq \mathcal{P} = \{p \mid \lenpath \leq n\}$ be the set of all paths of length up to $n$. Also, let  $\mathrm{P_{n}}$ denote the set of all path \emph{types} with length up to $n$, i.e. $\mathrm{P_{n}} = \{\typepath \mid p \in \mathcal{P}_n\}$. Let $\mathrm{P_{n}}(\ent_{1}, \rel) \subseteq \mathrm{P_{n}}$ denote all path types of length up to $n$ that originate at $\ent_{1}$ and end at the entities that are connected to $\ent_{1}$ by a direct edge of type $r$. In other words, if $S_{\ent_1 r} = \{\ent_{2} \mid (\ent_{1}, \rel, \ent_{2}) \in \mathcal{G}\}$ denotes the set of entities that are connected to $\ent_{1}$ via a direct edge $r$, then $\mathrm{P_{n}}(\ent_{1}, \rel)$ denotes the set of all path types of length up to n that start from $\ent_{1}$ and end at entities in $S_{\ent_1 r}$. By definition, $r \in \mathrm{P_{n}}(\ent_{1}, \rel)$. Similarly, we define $\mathcal{P}_n(\ent_{1}, \rel)$ which contain paths instead of path types.

\subsection{Model}
\label{sub:non_param_reasoning}

Given a query, our approach gathers KG path types from entities that are similar to the query entity. Each path type is weighed with respect to an estimate of both its frequency and precision (\S\ref{sub:contextual_entities}). By clustering similar entities together (\S\ref{sub:entity_clustering}), our model obtains robust estimate of the path statistics (\S\ref{sub:param_estimation}). Our approach is non-parametric because - (a) Instead of storing reasoning rules in parameters \cite{das2018go,minervini2019differentiable}, it derives them dynamically from $k$-similar entities (like  a non-parametric $k$-nn classifier \cite{cover1967nearest}). (b) We cluster entities together using a non-parametric clustering approach and provide an efficient way of adding / estimating parameters when entities are added to the KG (\S\ref{sub:open_world}).

\subsubsection{Reasoning from contextual entities}
\label{sub:contextual_entities}
Our approach first finds $k$ similar entities to the  query entity that have atleast an edge of type $r_q$. For example, for the query (\textsc{Melinda Gates}, \textsc{works\_in\_city}, ?), we would consider \textsc{Warren Buffet} if we observe (\textsc{Warren Buffet}, \textsc{works\_in\_city}, \textsc{Omaha}). We refer to these entities as `contextual entities'.  Each entity is represented as a sparse vector of its outgoing edge types, i.e. $\mathbf{e_i} \in \{0,1\}^{|\mathcal{R}|}$. If entity $e_i$ has $m$ distinct outgoing edge types, then the dimension corresponding to those types are set to 1. This is an extremely simple and flexible way of representing entities which we find to work well. Also note that, as more data is added about an entity, this sparse representation makes it trivial to update the embeddings.

Let $E_{c,q}$ denote the set of contextual entities for the query $q$. To compute $E_{c,q}$, we first sort entities with respect to their cosine distance with respect to query entity and select the $k$ entities with the least distance and which have the query relation $r_q$. For each contextual entity $e_c$, we gather the path types (up to length n) that connect $e_c$ to the entities it is connected by the edge $r_q$ (i.e. $\mathrm{P_n}(e_c, r_q)$ in \S\ref{sub:notation}). These extracted path types will be used to reason about the query entity. Let $\mathrm{P_n}(E_{c,q}, r_q) = \bigcup_{e_c \in E_{c,q}} \mathrm{P_n}(e_c, r_q)$ represent the set of unique path types from the contextual entities. The probability of finding the answer entity $\ent_{2}$ given the query is given by:
\begin{align}
    P\left(\ent_{2} \mid e_{1q}, r_q\right) =  \sum_{p \in \mathrm{P_n}(E_{(c,q)}, r_q)} P(\ent_{2}, p \mid  e_{1q}, r_q) \nonumber \\ 
 =  \sum_{p} P(p \mid e_{1q}, r_q) P(\ent_{2} \mid p, e_{1q}, r_q)
 \label{eq:factorization}
\end{align}

We marginalize the random variable representing the path types obtained from $E_{c,q}$. $P(p \mid e_{1q}, r_q)$ denotes the probability of finding a path type given the query. This term captures how frequently each path type co-occurs with a query and represents the prior probability for a path type. On the other hand,  $P(\ent_{2} \mid p, e_{1q}, r_q)$  captures the proportion of times, when a path type $p$ is traversed starting from the query entity, we reach the correct answer instead of some other entity. This term can be understood as capturing the likelihood of reaching the right answer or the 'precision' of a reasoning path type. This is crucial in penalizing `spurious' path types that sometimes coincidentally find the right answer entity. For example, for the query relation \textsc{works\_in\_city}, the path type (\textsc{Friend}$\wedge$ \textsc{Lives\_in\_City}) might have a high prior probability (since people often have many friends in the city where they work). However, this path is `spurious' with respect to \textsc{works\_in\_city}, since they might have friends living in various cities and hence this path type will not necessarily return the correct answer.

\begin{figure*}
    \centering
    \includegraphics[width=2\columnwidth]{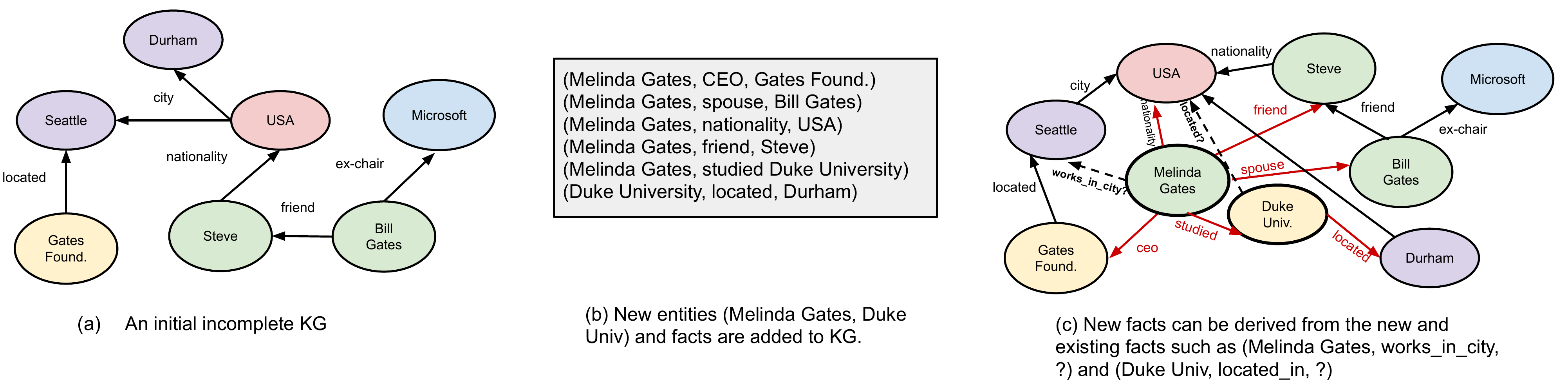}
    \caption{We consider a setting where \emph{new entities} and facts are added continuously to the KG. Our non-parametric approach can seamlessly reason with the newly added entities and can infer new facts about them (e.g. (\textsc{Melinda}, \textsc{Works\_in\_City}, ?) or (\textsc{Duke Univ.}, \textsc{located\_in\_country}, ?)) without requiring expensive training.}
    \label{fig:open_world}
\end{figure*}

\subsubsection{Entity Clustering}
\label{sub:entity_clustering}
Equation~\ref{eq:factorization} has parameters for each entity in the KG. For large KGs, this can quickly lead to parameter explosion. Also, estimating per-entity parameter leads to noisy estimates due to sparsity. Instead, we choose to cluster similar entities together. Let $c$ be a random variable representing the cluster assignment of the query entity. Then for the path-prior term, we have 
\begin{align*}
    P(p \mid e_{1q}, r_q) = \sum_{c} P(c \mid e_{1q}, r_q) P(p \mid c, e_{1q}, r_q)
\end{align*}

We assume that each entity is assigned to one cluster, so $ P(c \mid e_{1q}, r_q)$ is zero for all clusters except the cluster in which the query entity belongs to. Secondly we assume, that the prior probability of a path given the entity and cluster can be determined from the cluster alone and is independent of each entity in the cluster. In other words, if $c_{e_{1q}}$ is the cluster in which the $e_{1,q}$ has been assigned, then $P(p \mid c_{e_{1q}}, e_{1q}, r_q) = P(p \mid c_{e_{1q}}, r_q)$. Instead of per-entity parameters, we now aggregate statistics over entities in the same cluster and have per-cluster parameters. We also show that this leads to significantly better performance (\S\ref{sub:kbc}). A similar argument applies for the path-precision term in which we calculate the proportion of times, a path leads to the correct answer entity starting from each entity in the cluster.

To perform clustering, we use hierarchical agglomerative clustering with average linkage with the entity-entity similarity defined in \S \ref{sub:contextual_entities}. 
We extract a non-parameteric number of clusters from the hierarchy using a threshold on the linkage function. Agglomerative clustering has been shown to be effective in many knowledge-base related 
tasks such as entity resolution \cite{lee-etal-2012-joint,vashishth2018cesi} and in general has shown to 
outperform flat clustering methods such as K-means \cite{green2012entity,kobren2017hierarchical}. A flat clustering is extracted from the hierarchical clustering by using a threshold on the linkage function score. We perform a breadth first search from the root of the tree stopping at nodes for which the linkage is above the given threshold. The nodes where the search stops give a flat clustering (refer to \S\ref{sub:appendix_flat_cluster} for more detail on this).

\subsubsection{Parameter Estimation}
\label{sub:param_estimation}
Next we discuss how to estimate path prior and precision terms. There exists abundant modeling choices to estimate them. For example, following \citet{chen2018variational}, we could train a neural network model to estimate $P(p \mid c_{e_{1q}}, r_q)$. However, with our original goal of designing a simple and efficient non-parametric model, we estimate these parameters by simple count statistics from the KG. E.g., the path prior $P(p \mid c, r_q)$ is estimated as
\begin{align}
     \frac{\sum_{e_c \in c} \sum_{p' \in \mathcal{P}_{n}(e_c, r_q)}\mathbbm{1}\left[\mathsf{type}(p') = p\right]}{\sum_{e_c \in c} \sum_{p' \in \mathcal{P}_{n}(e_c, r_q)}\mathbbm{1}}
    \label{eq:prior}
\end{align}
For each entity in cluster $c$, we consider the paths that connect $e_c$ to entities it is directly connected to via edge type $r_q$ ($ \mathcal{P}_{n}(e_c, r_q)$ in \S\ref{sub:notation}). The path prior for a path type $p$ is computed as the proportion of times the  type of paths in  $\mathcal{P}_{n}(e_c, r_q)$ is equal to $p$. Note that in equation \ref{eq:prior}, if a path type appears multiple times, we count all instances. For example, for the query relation \textsc{Works\_in\_City}, a path of the form (\textsc{Co\_worker} $\wedge$ \textsc{Works\_in\_City}) can occur multiple times, since a person can have multiple different co-workers. Considering just path types will lead to under-weighing of such important paths. Similarly, the path-precision probability ($P(\ent_{2} \mid p, c, r_q)$) can be estimated as,

\begin{align}
& \frac{\sum_{e_c \in c}\sum_{p' \in \mathcal{P}_{n}(e_c)} \mathbbm{1}[\mathsf{type}(p') = p] \cdot \mathbbm{1}[\mathrm{en}(p') \in S_{\ent_c r_q}]}{\sum_{e_c \in c}\sum_{p' \in \mathcal{P}_{n}(e_c)} \mathbbm{1}[\mathsf{type}(p') = p]}
\label{eq:precision}
\end{align}
Let $\mathcal{P}_{n}(e_c)$ denote the paths of up to length $n$ starting from the entity $e_c$. Note, unlike $\mathcal{P}_{n}(e_c, r_q)$, the paths in $\mathcal{P}_{n}(e_c)$ do not have to end at specific entities. Also from \S\ref{sub:notation}, $\mathrm{en}(p)$ denotes the end entity for a path $p$ and $S_{\ent_c r_q}$ denotes the set of entities that are connected to $e_c$ via a direct edge of type $r_q$. Equation \ref{eq:precision}, therefore, estimates the proportion of times the path $p$ successfully ends at one of the answer entities when starting from $e_c$, given $r_q$. 

There are several advantages in estimating the parameters using simple count statistics. Firstly, they are extremely simple, and statistics for each entity in clusters can be computed in parallel making them extremely time efficient. Secondly once they are computed, our approach needs \emph{no further training}. Lastly, when new data is added, it makes it easy to update the parameters without training from scratch. 

To summarize, given a query entity $(e_{1q}, r_q)$, our method gathers reasoning paths from $k$ similar entities to $e_{1q}$. These reasoning paths are then traversed in the KG starting from $e_{1q}$, leading to a set of candidate answer entities. The score of each answer entity candidate is computed as a weighted sum of the reasoning paths the lead to them (Equation~\ref{eq:factorization}). Each path is weighed with an estimate of its frequency (Equation~\ref{eq:prior}) and precision (Equation~\ref{eq:precision}) given the query relation. The next section describes how we extend our model for open-world setting where new entities and facts are added to the KB.

\subsection{Open-world Setting}
\label{sub:open_world}
A great benefit of non-parametric models is that it can seamlessly handle growing data by adding new parameters. New entities constantly arrive in the world (e.g. new Wikipedia articles about entities are frequently created). We consider a setting (Figure~\ref{fig:open_world}) in which new entities with few facts (edges) about them keep getting added to the KG. This setting is challenging for parametric models \cite{das2018go,sun2019rotate} as it is unclear how these models can incorporate new entities without retraining from scratch. However, retraining to obtain entity embeddings on industrial scale KGs might be impractical (e.g. consider Facebook social graph where new users are joining continuously). Next, we show that our approach can handle this setting efficiently in the following way:\\
(a) \textbf{Adding/updating entity representations}: First we need to create entity representations for the newly arrived entities. Also, for some existing entities for which new edges were added (e.g. \textsc{Bill Gates}, \textsc{Durham}, etc. in figure~\ref{fig:open_world}), their representations need to be updated. Recall, that we represent entities as a sparse vector of its edge types and hence this step is trivial for our approach.\\
(b) \textbf{Updating cluster assignments}: Next the new entities needs to be added to clusters of similar entities. Also, the cluster assingments of entities that got updated can also change as well and their change can further trigger changes to the clustering of other entities. To handle this, one could naively cluster all entities in the KG, however that could be wasteful and time-consuming for large KGs. Instead, we use an online hierarchical clustering algorithm - \textsc{Grinch} \cite{monath2019scalable}, which has shown to perform as well as agglomerative clustering in the online setting. \textsc{Grinch} observes one entity at a time, placing it next to its nearest neighbor and performing local re-arrangements in the form of rotations of tree nodes and global re-arrangments in the form of grafting a subtrees from part of the tree to another. Entities can be deleted from a hierarchy by simply removing the corresponding leaf node. We first use \grinch to delete the entities whose representations had changed because of the addition of the new node and then incrementally add those entities back along with the newly added entities in the KG. We extract a flat clustering from the hierarchical clustering built by \textsc{Grinch} using the same method as in \S\ref{sub:entity_clustering}.\\
(c) \textbf{Re-estimating new parameters}: After re-assigning clusters, the final step is to estimate the per-cluster parameters. This computation is efficient as it is clear from equations \ref{eq:prior} and \ref{eq:precision} that the contribution from each entity in a cluster can be computed independently (and hence can be easily parallelized). However, even for each entity, this computation needs path traversal in the KG which is expensive. We show that we do not have to re-compute for all entities in the clusters.

Let $n$ denote the maximum length of a reasoning path considered by our model. For every new entity $e_i$ added to the KG, we need to recompute statistics for entities that lie within cycles of length up to $(n+1)$ starting from $e_i$. 
Please refer to appendix (\ref{sub:appendix_estimate_params}) for a justification of this result.
\section{Experiments}
\label{sec:experiments}
\begin{table}[]
    \centering
    \begin{tabular}{cccc}
    \toprule
        &  $| \mathcal{V} |$ & $| \mathcal{R} |$ & $|E|$ \\
        \midrule
         NELL-995 & 75,492 & 200 & 154,213  \\
         FB122 & 9,738 & 122 & 112,476 \\
         WN18RR & 40,943 & 11 & 93,003 \\
         \bottomrule
    \end{tabular}
    \caption{Dataset Statistics}
    \label{tab:dataset_stats}
\end{table}

\begin{table*}
\centering

\resizebox{\textwidth}{!}{
\begin{tabular}{clcccccccccccccc}
\toprule
& & \multicolumn{4}{c}{Test-I} & & \multicolumn{4}{c}{Test-II} & & \multicolumn{4}{c}{Test-ALL}\\
\cline{3-6}\cline{8-11}\cline{13-15}
& & \multicolumn{3}{c}{Hits@N (\%)} & \multirow{2}{*}{MRR} & & \multicolumn{3}{c}{Hits@N (\%)} & \multirow{2}{*}{MRR} & & \multicolumn{3}{c}{Hits@N (\%)} & \multirow{2}{*}{MRR}
\\
\cline{3-5}\cline{8-10}\cline{13-15}
& & 3 & 5 & 10 & & \ & 3 & 5 & 10 & \ & & 3 & 5 & 10 & \\
\toprule
\multirow{5}{*}{\rotatebox[origin=c]{90}{\parbox[c]{1.5cm}{\centering {\bf With Rules}}}}
& KALE-Pre~\citep{guo2016jointly} & 35.8 & 41.9 & 49.8 & 0.291 & & 82.9 & 86.1 & 89.9 & 0.713 & & 61.7 & 66.2 & 71.8 & 0.523 \\
& KALE-Joint~\citep{guo2016jointly} & {38.4} & {44.7} & {52.2} & 0.325 & & 79.7 & 84.1 & 89.6 & 0.684 & & 61.2 & 66.4 & 72.8 & 0.523 \\
& \emph{ASR}-DistMult~\citep{minervini2017adversarial} & 36.3 & 40.3 & 44.9 & 0.330 & & 98.0 & 99.0 & 99.2 & 0.948 & & 70.7 & 73.1 & 75.2 & 0.675 \\
& \emph{ASR}-ComplEx~\citep{minervini2017adversarial} & 37.3 & 41.0 & 45.9 & { 0.338} & & {\bf 99.2} & {\bf 99.3} & {\bf 99.4} & {\bf 0.984} & & 71.7 & 73.6 & 75.7 & 0.698 \\
& \textsc{KBlr}~\citep{garcia2017kblrn} & -- & -- & -- & -- & & -- & -- & -- & -- & & {74.0} & \textbf{77.0} & \textbf{79.7} & {0.702} \\
\midrule
\multirow{8}{*}{\rotatebox[origin=c]{90}{\parbox[c]{1.5cm}{\centering {\bf Without Rules}}}}
& TransE~\citep{bordes2013translating} & 36.0 & 41.5 & 48.1 & 0.296 & & 77.5 & 82.8 & 88.4 & 0.630 & & 58.9 & 64.2 & 70.2 & 0.480 \\
& DistMult~\citep{distmul} & 36.0 & 40.3 & 45.3 & 0.313 & &  92.3 & 93.8 & 94.7 & 0.874 & & 67.4 & 70.2 & 72.9 & 0.628 \\
& ComplEx~\citep{trouillon2016complex} & 37.0 &  41.3 & 46.2 &  0.329 & & 91.4 & 91.9 & 92.4 & 0.887 & & 67.3 & 69.5 & 71.9 & 0.641 \\
& GNTPs~\citep{minervini2019differentiable} & 33.7 & 36.9 & 41.2 & 0.313 & & {\bf 98.2} & {\bf 99.0} & {\bf 99.3} & {\bf 0.977} & & 69.2 &  71.1 & 73.2 & 0.678 \\
& RotatE \citep{sun2019rotate} & \textbf{51.1} & \textbf{55.1} & \textbf{60.3} & \textbf{0.471} & & 86.8 & 88.6 & 90.7 & 0.846 & & 70.8 & 73.57 & 77.0 & 0.678 \\
& CBR~\citep{cbr} & 40.0 & 44.5 & 48.8 & 0.359 & & 67.8 & 71.8 & 75.9 & 0.636 & & 57.0 & 61.2 & 65.3 & 0.527 \\
& Our Model & 49.0 & 52.7 & 57.1 & 0.457 & & 94.8 & 95.0 & 95.3 & 0.948 & & \textbf{74.2} & \textbf{76.0} & \textbf{78.2} & \textbf{0.727} \\
\bottomrule
\end{tabular}
}
\caption{Link prediction results on \fb. Test-II denotes a subset of triples that can be inferred via logical rules.} \label{tab:fb122}
\end{table*}

\begin{table*}
\centering
\small
%\vspace{-2mm}
\begin{tabular}{@{} l  c c c c c c c c c @{}}\toprule
 \textbf{Metric} & \textbf{TransE} & \textbf{DistMult} & \textbf{ComplEx}  & \textbf{ConvE} & \textbf{RotatE} & \textbf{GNTP} & \textbf{MINERVA} & \textbf{\cbr} & \textbf{Our Model}\\\midrule
\textsc{hits}@1  & - & 0.39 & 0.41 & 0.40 & \textbf{0.43} & 0.41 & 0.40 &     0.38 & \textbf{0.43}\\
 \textsc{hits}@3  & - & 0.44 & 0.46 & 0.44 & \textbf{0.49} & 0.44 & 0.43 & 0.46 & \textbf{0.49}\\
 \textsc{hits}@10 & 0.50 & 0.49 & 0.51 & 0.52 & \textbf{0.57} & 0.48 & 0.49 & 0.51 & 0.55\\
 \textsc{MRR}     & 0.23 & 0.43 & 0.44 & 0.43  & \textbf{0.48} & 0.43 & 0.43 & 0.43 & \textbf{0.48}\\
\midrule
\textsc{hits}@1 & 0.53 & 0.61 & 0.61 & 0.67 & 0.65 & - & 0.66 & 0.70&  \textbf{0.77} \\
\textsc{hits}@3 & 0.79 & 0.73 & 0.76 & 0.81 & 0.82 & - & 0.77 & 0.83 & \textbf{0.85}\\
\textsc{hits}@10 & 0.87 & 0.79 & 0.83 & 0.86 & 0.87 & - & 0.83 & 0.87 & \textbf{0.89}\\
\textsc{MRR} & 0.67 & 0.68 & 0.69 & 0.75 & 0.74 & - & 0.72  & 0.77 & \textbf{0.81}\\
\bottomrule
\end{tabular}
% \vspace{-2mm}
\caption{Results on \wn (above) and \nell (tail-prediction;below)}
\label{tab:wn_nell}
\vspace{-2mm}
\end{table*}

In this section, we evaluate our proposed approach on a wide array of 
knowledge-base completion (KBC) benchmarks (\S \ref{sub:kbc}). To evaluate the non-parametric nature of our approach, we also evaluate on an `open-world' setting (\S\ref{sub:open_world}) in which new entities are added to the KG. We demonstrate our proposed approach is competitive to several 
state-of-the-art methods on benchmarks in the standard setting, but it greatly outperforms
other methods in the online setting (\S\ref{sub:online}). The best hyper-parameters for all experiments including the range of hyper-parameter tried and results on validation set are noted in \S\ref{sub:repro_check}.

\subsection{Data and Evaluation Protocol}
\label{sub:data_and_eval_protocols}

\textbf{Data}. We evaluate on the following KBC datasets: \textbf{NELL-995}, \textbf{FB122} \cite{guo2016jointly}, \textbf{WN18RR} \cite{dettmers2018convolutional}. \textbf{FB122} is a subset of the dataset derived from Freebase, FB15K \cite{bordes2013translating}, containing 122 relations regarding people, locations, and sports. {NELL-995} \cite{deeppath} a subset of the NELL derived from the 995th iteration of the system. WN18RR was created by \citet{dettmers2018convolutional} from WN18 by removing inverse relation test-leakage.

\textbf{Evaluation metrics}. Following previous work, we evaluate our method using \textsc{hits}@N and mean reciprocal rank (MRR), which are standard metrics for evaluating a ranked list.

\subsection{Experimental Setting}
\label{sub:experimental_setting}

\textbf{Knowledge Base Completion}. Given an entity $\ent_1$ and a relation $\rel$, our task is retrieve all entities $\ent_2$ such that $(\ent_1, \rel, \ent_2)$ belongs in the edges $E$ in a KG $\mathcal{G}$. This task is known as \emph{tail prediction}. If the relation is instead the inverse relation $\rel^{-1}$, we assume that we are given an  $\ent_2'$ and asked to predict entities $\ent_1'$ such that $(\ent_1', \rel^{-1}, \ent_2')$ belongs in the edges $E$ (\emph{head prediction}). To be exactly comparable to baselines, we report an average of head and tail prediction results\footnote{except for \nell dataset where like our baselines, we report tail-prediction performance.}. We are given a knowledge graph with three partitions of edges, $E_\text{train}$, $E_\text{dev}$, $E_\text{test}$.

For this task, we evaluate against several state-of-the-art embeddings based models such as 
DistMult \cite{distmul}, ComplEx \cite{trouillon2016complex}, ConvE \cite{dettmers2018convolutional}, RotatE \cite{sun2019rotate}. We also compare against several parametric rule learning methods --- 
NTP \cite{ntp}, NeuralLP \cite{yang2017differentiable}, MINERVA \cite{das2018go}, GNTP \citep{minervini2019differentiable} and also the closely related CBR approach of \citet{cbr}.

\noindent \textbf{Open-world Knowledge Base Completion}. In this setting, we begin with the top 10\% of the most popular nodes (with several edges going out from them) and add more randomly selected nodes such that the initial seed KB contains 50\% of all the entities in $\mathcal{V}$. This is to ensure, that the seed KB is not too sparse and the initial models trained on them are meaningful. Next, any edges between the nodes selected are added to the seed KB. We divide the rest of the entities randomly into 10 batches. Each batch of entities is incrementally added to the KB along with the edges contained in it. The validation and test set are also divided in the same way, i.e. if both the head and tail entity of a triple are present in the KB, only then the triple is put in the corresponding splits.

Parametric models for KBC that learn representations for a fixed set of entities can not handle `open-world' setting out-of-the-box. We extend the most competitive embedding based model - RotatE \cite{sun2019rotate} for this task. For every new entity arriving in a batch, we initialize a new entity embedding for it. We explore two ways of initializing the new entity embeddings --- (a) random initialization, and (b) average of element-wise rotation of entity embeddings w.r.t the relation that this new entity is connected to. Specifically, let $t$ denote the new entity and let $S = \{(h, r, t)\}$ be the facts associated with entity $t$. Then the embedding $\mathbf{e_t}$ is computed as
\begin{equation}
\mathbf{e_{t}}= \frac{\sum_{(h, r, t) \in S}\, \mathbf{e_h}\circ \mathbf{e_r}}{|S|}
\end{equation}
Here, $\circ$ represents the Hadamard (or element-wise) product. This initialization minimizes the RotatE objective for the new embedding ensuring that it is ``well-placed'' according to the model in the previous time step. Embeddings for new relations are initialized randomly. Next, the model is further trained on the new batch of triples so that the new entity embeddings get trained.  Note, for massive KGs, it might be impractical to re-train on the entire data as new batches of data arrive frequently, however to still prevent the model to forget what it had learned before, we also sample $m$\% of triples that it had already been trained on and re-train on them. We ensure that triples in the neighborhood of the newly added entities are ten times likely to be sampled more than other triples. We also try a setting where we try freezing the initially trained entity embeddings and only training the new entity and relation embeddings.

\subsection{Results on KBC benchmarks}
\label{sub:kbc}
The results for KBC tasks are presented in Table~\ref{tab:fb122} and \ref{tab:wn_nell}\footnote{There are no reported results of GNTPs on \nell}. Our method does significantly better than parametric rule learning approaches such as MINERVA, GNTPs and the recent case-based approach of \citet{cbr}. We would like to highlight the difference between the performance of our model and that of \citet{cbr} on the test-II evaluation of \fb where triples can be answered by learning logical rules. This results emphasizes the importance of our probabilistic weighing of paths. We also perform comparably to most embedding based models and achieve state-of-the-art results on the overall test sets of \fb and \nell. We report the mean over 3 runs for our model.

We perform an ablation where we do not cluster entities (i.e. every entity has its own cluster) and have per-entity parameters. 
Table~\ref{tab:wn18rr_clustering} notes the drop in performance due to the noisy estimates of path prior and precision parameters because of sparsity. Table \ref{tab:nell_eye_candy} shows an example where our model learns to score different paths based on the type of entities present in the cluster.

\textbf{Effect of path length on \wn:} On the dev set of WN18RR, out of $2985$ queries where our method does not rank the answer in the top-10, $2030$ queries require a minimum path length greater than 3. Path-based reasoning models have no power to answer these queries. To correct for this, we perform an experiment with the path length $n=5$ ($950$ of $2030$ answers are reachable). The results in Table \ref{tab:wn18rr_pathlen} show that our method recovers a significant portion of performance when allowed to use longer reasoning paths.

\begin{table}
\centering
\small
%\vspace{-2mm}
\begin{tabular}{ l  c c}
\toprule
& \textbf{Our Method} & \textbf{Our Method}   \\ 
&  & \bf w/o clustering   \\
\midrule
  \textsc{hits}@1 & 0.42 & 0.29\\
  \textsc{hits}@3  & 0.46 & 0.36\\
  \textsc{hits}@10  & 0.51 & 0.45\\
  \textsc{MRR}   & 0.45 & 0.34\\
\bottomrule
\end{tabular}
\vspace{-2mm}
\caption{Impact of clustering on WN18RR}
\label{tab:wn18rr_clustering}
% \vspace{-8mm}
\end{table}

\begin{table}
\centering
\small
%\vspace{-2mm}
\begin{tabular}{ l c c c}
\toprule
& \textbf{RotatE} & \textbf{Our Method} & \textbf{Our Method}   \\
& & \textbf{($n=3$)} & \textbf{($n=5$)}   \\
\midrule
  \textsc{hits}@1 & 0.43 & 0.42 & 0.43\\
  \textsc{hits}@3  & 0.49 & 0.46 & 0.49\\
  \textsc{hits}@10  & 0.57 & 0.51 & 0.55\\
  \textsc{MRR}   & 0.48 & 0.45 & 0.48\\
\bottomrule
\end{tabular}
\vspace{-2mm}
\caption{Impact of path length on WN18RR}
\label{tab:wn18rr_pathlen}
\vspace{-4mm}
\end{table}

\begin{figure*}
    \centering
    \includegraphics[width=0.24\textwidth]{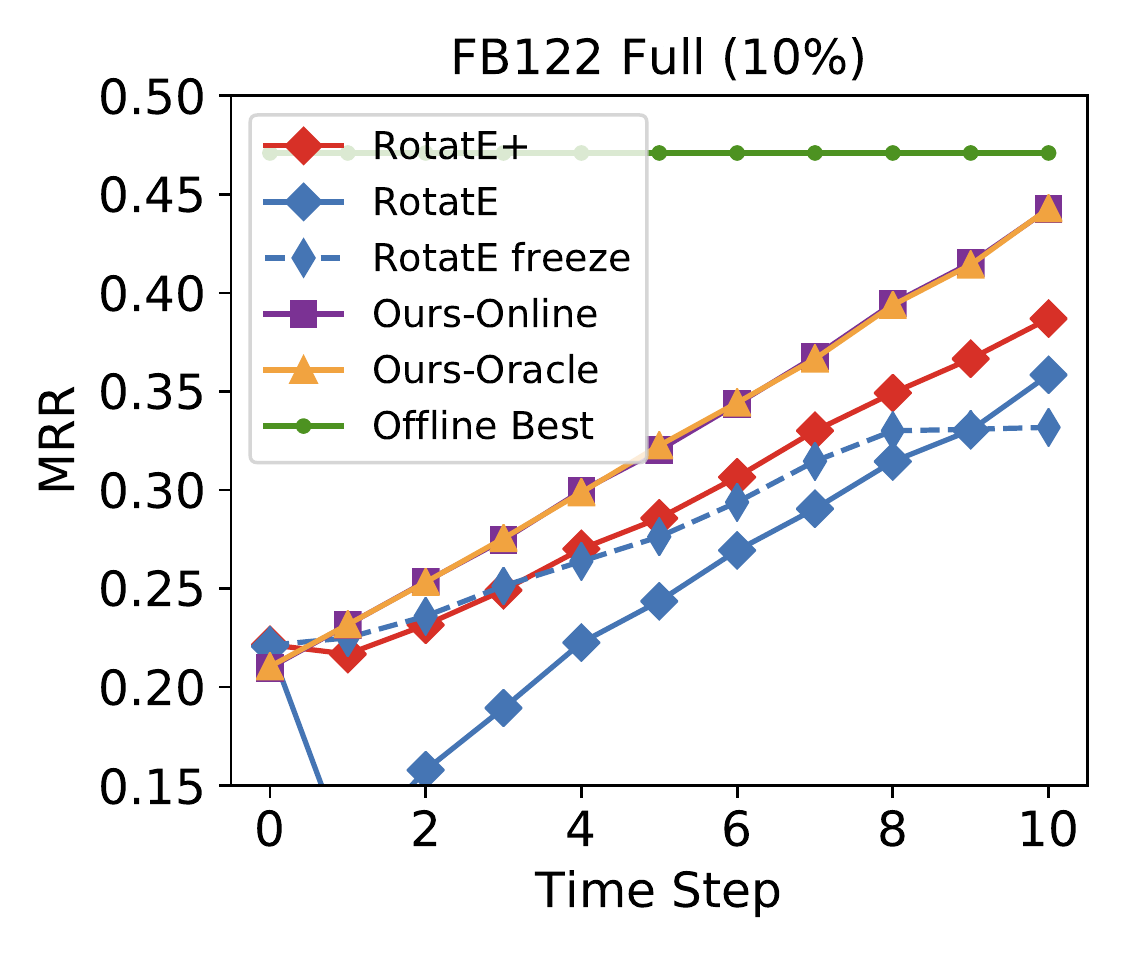}
    \includegraphics[width=0.24\textwidth]{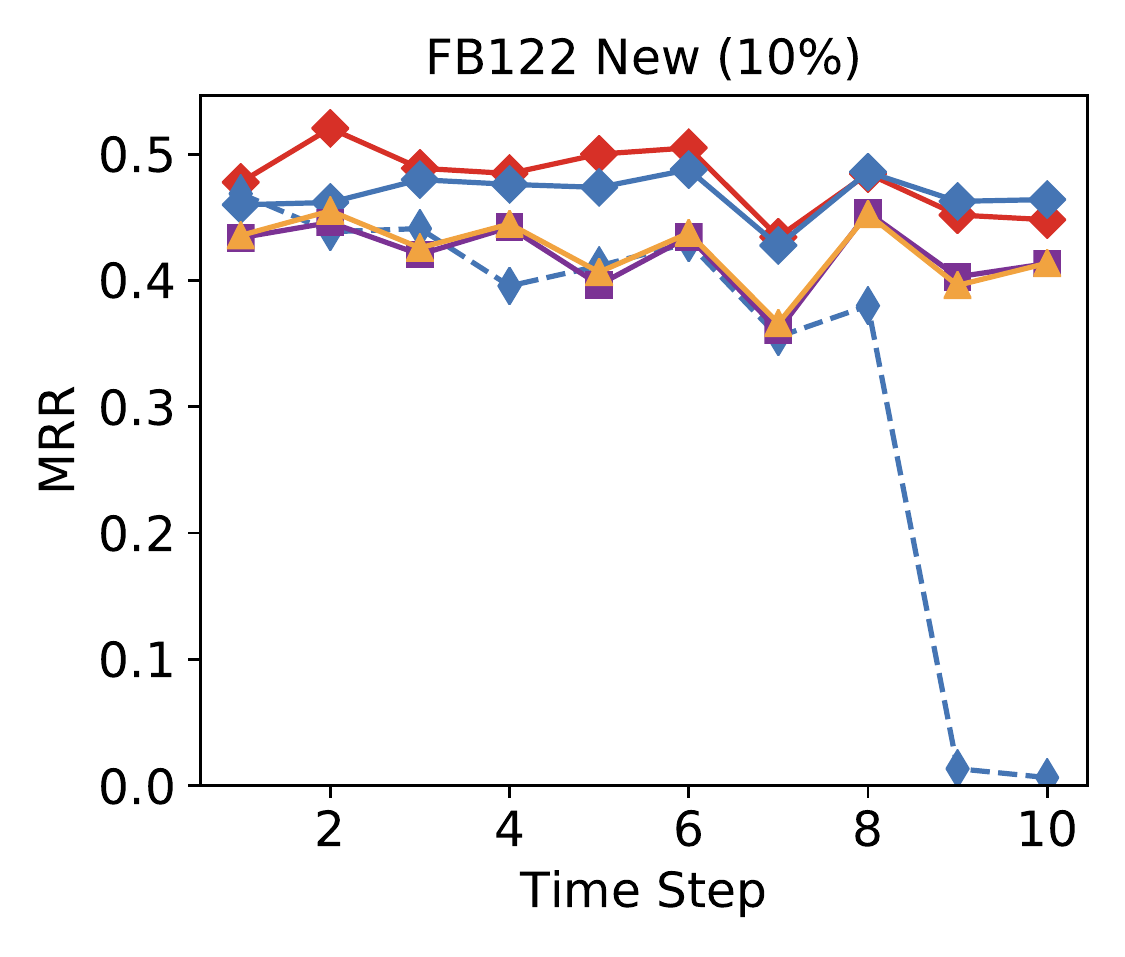}
    \includegraphics[width=0.24\textwidth]{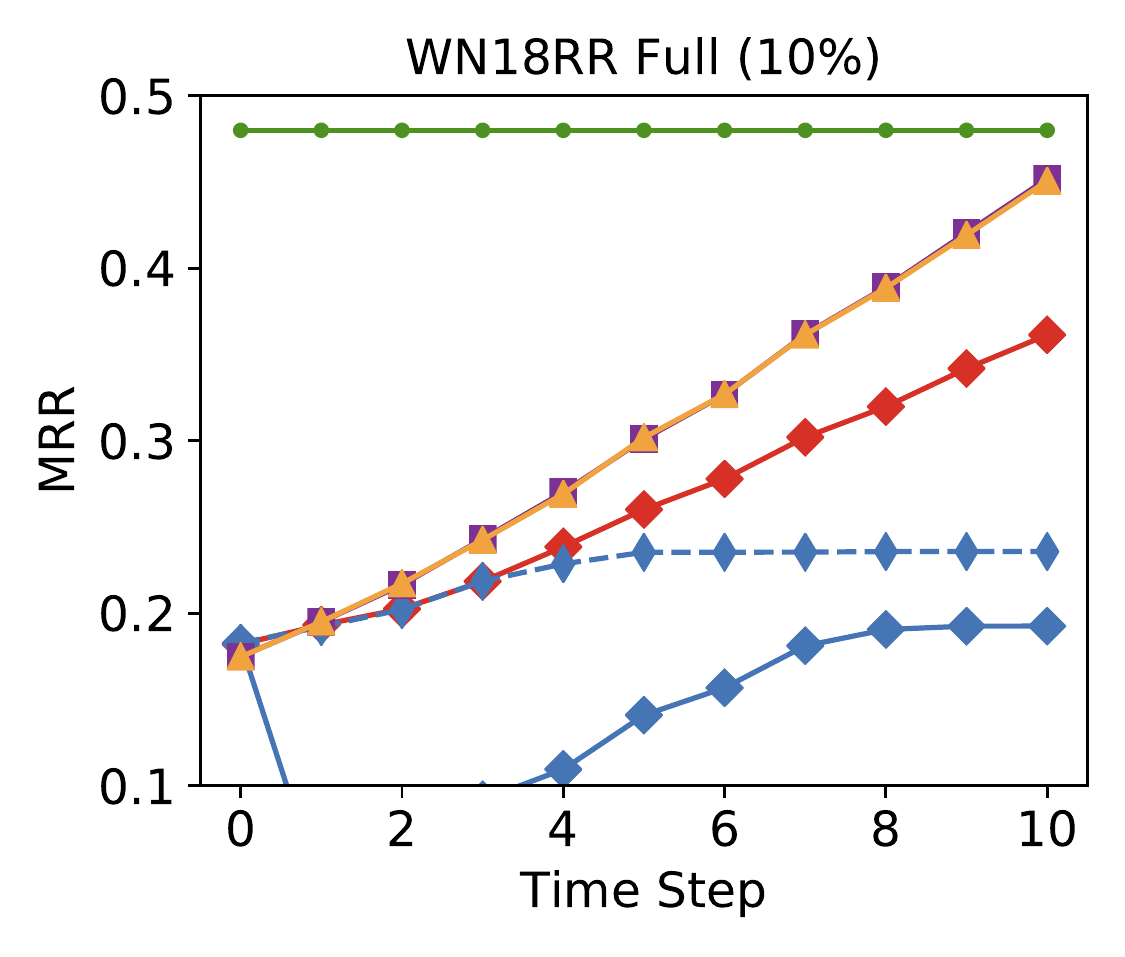}
    \includegraphics[width=0.24\textwidth]{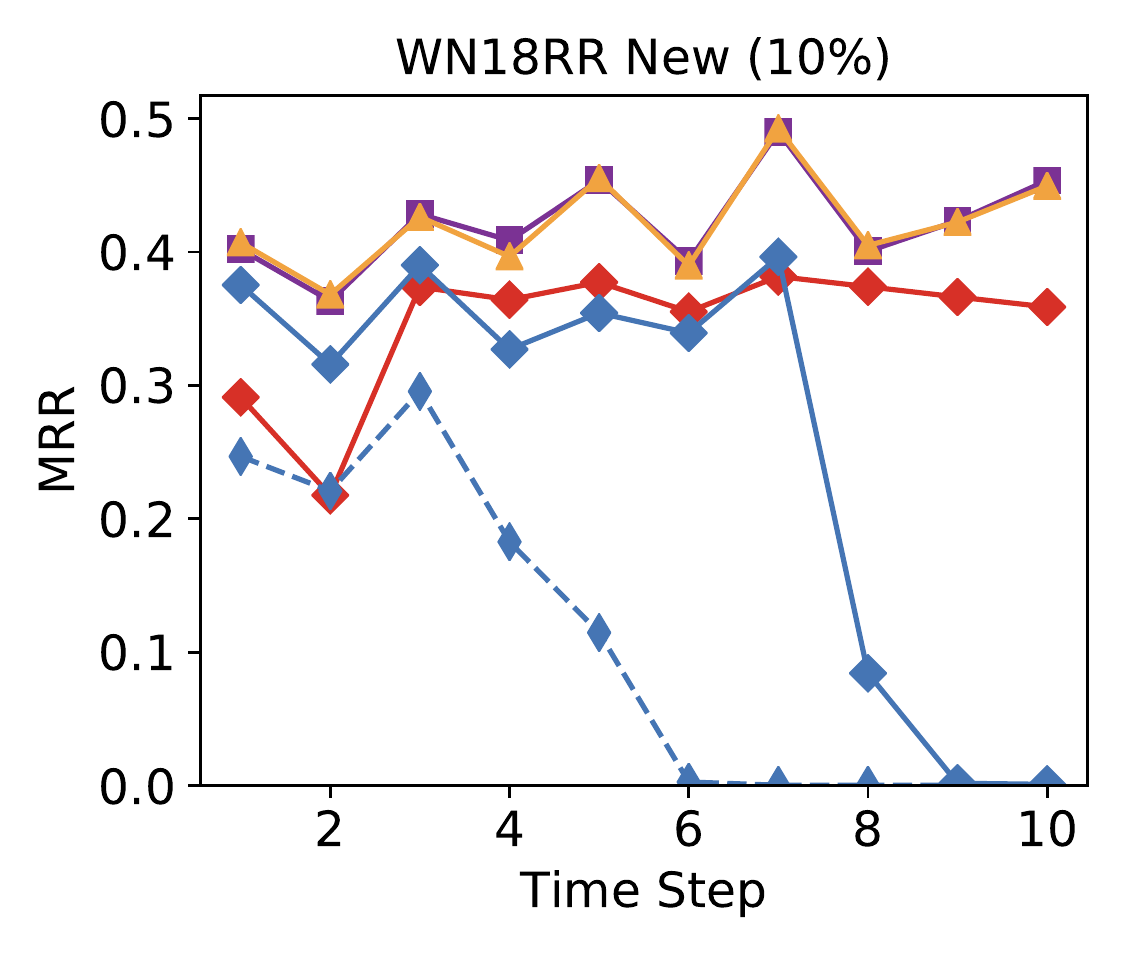}
    \includegraphics[width=0.24\textwidth]{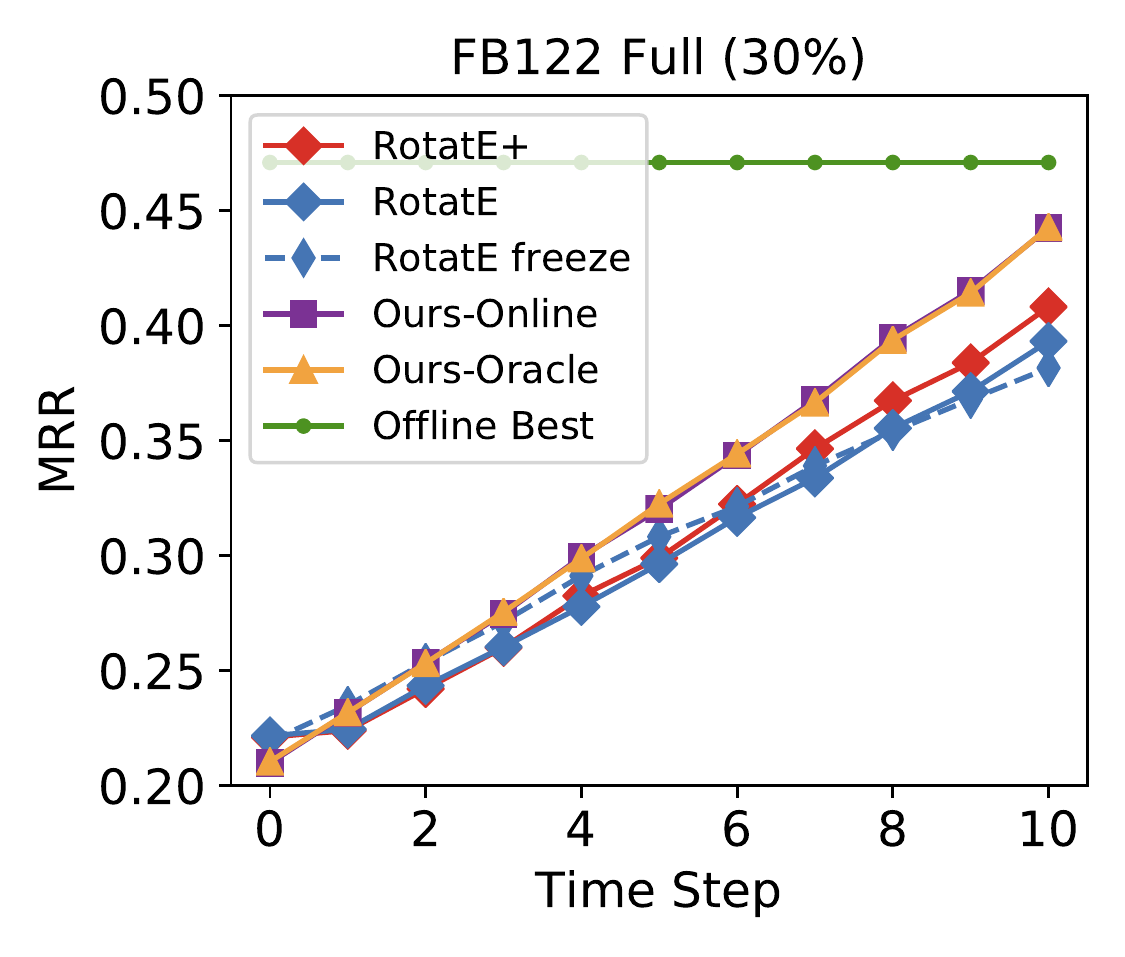}
    \includegraphics[width=0.24\textwidth]{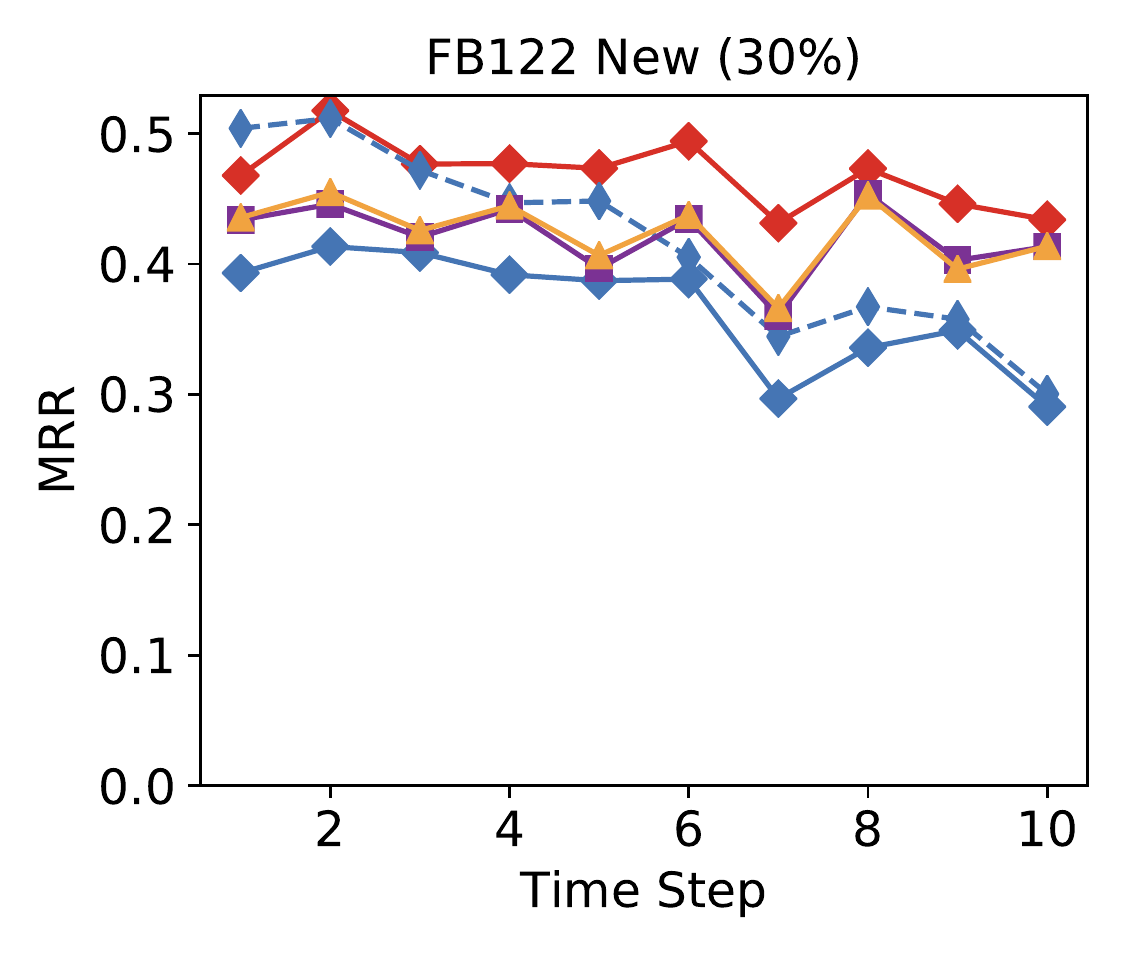}
    \includegraphics[width=0.24\textwidth]{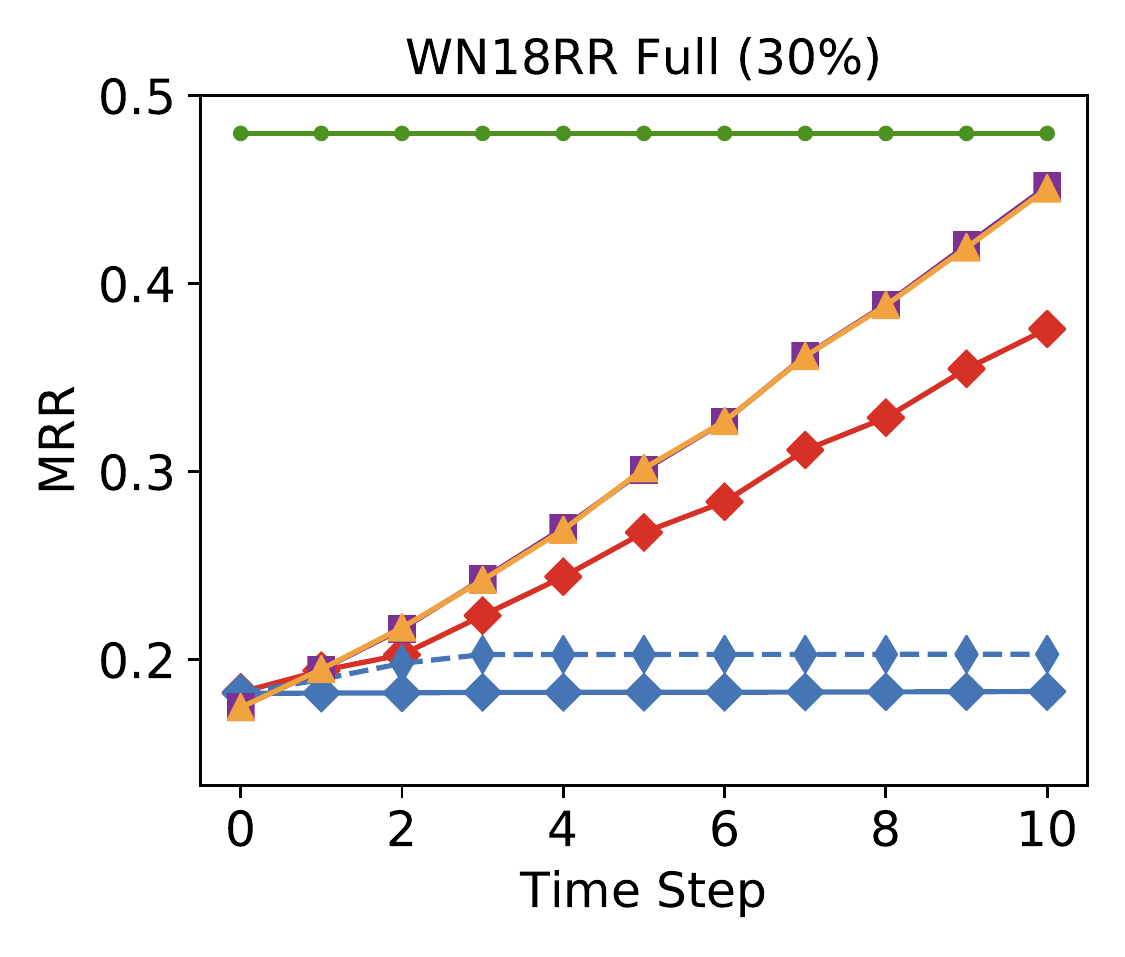}
    \includegraphics[width=0.24\textwidth]{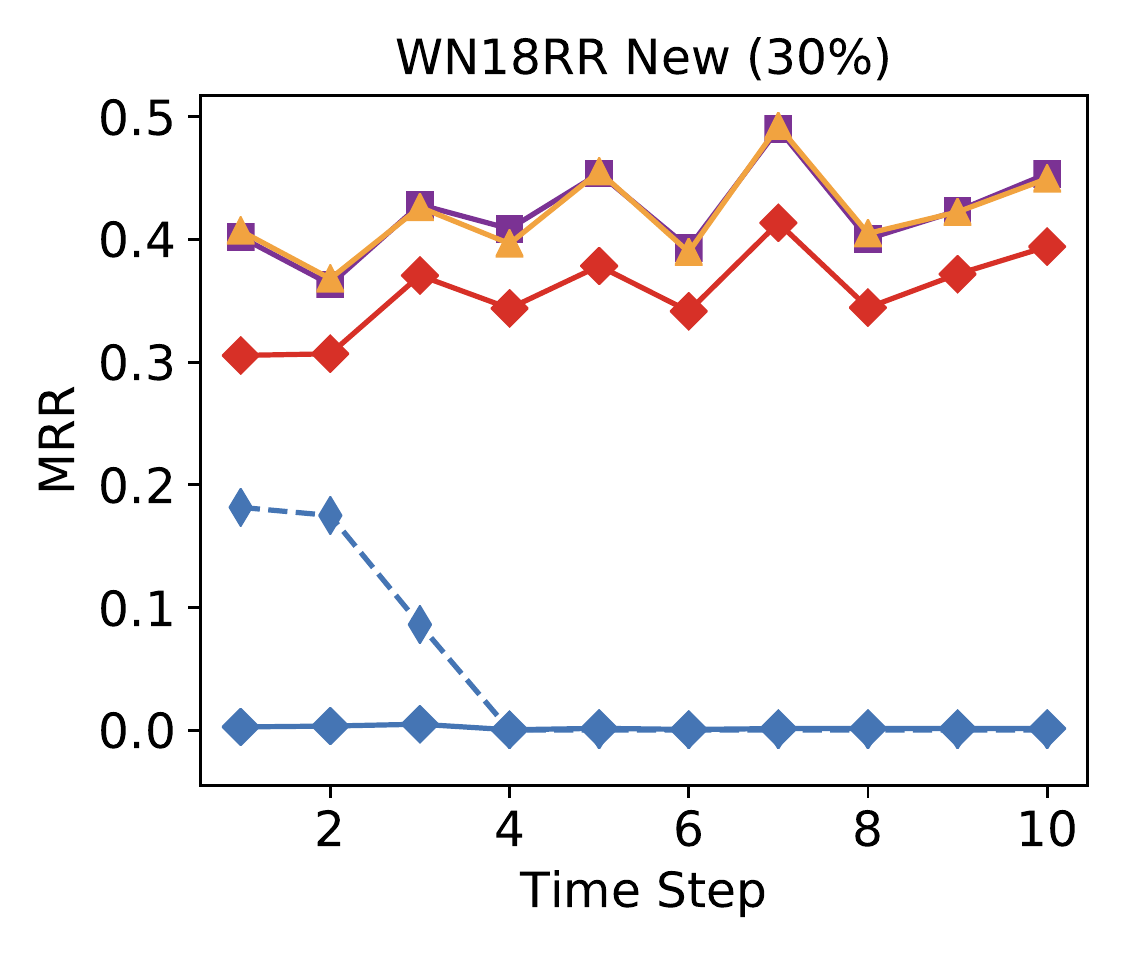}
    \caption{Results for open-world setting when trained with 10\% (top row) and 30\% (bottom row) of already seen edges. Our online method matches the offline version of our approach and outperforms the online variants of RotatE. After all data is observed our online method achieves results closest to the best offline method's results.}
    \label{fig:streaming}
\end{figure*}

\begin{table*}
    \centering
    \small
    \begin{tabular}{c c}
    \toprule
      \multirow{2}{*}{\bf Athlete Cluster}  & (\texttt{athlete-led-sports-team}, \texttt{team-plays-in-league})\\
         & (\texttt{athlete-home-stadium}, \texttt{league-stadiums}$^{-1}$)\\\midrule
        \multirow{3}{*}{\bf Politician Cluster}  & (\texttt{politician-us-member-of-political-group}, \texttt{person-belongs-to-organization}$^{-1}$, \\
        & \texttt{agent-belongs-to-organization}) \\
         & (\texttt{agent-collaborates-with-agent}, \texttt{agent-belongs-to-organization})\\\bottomrule
    \end{tabular}
    \caption{High scoring paths in different clusters for the query \texttt{agent-belongs-to-organization} in \nell}
    \label{tab:nell_eye_candy}
\end{table*}

\subsection{Open-World KBC results}
\label{sub:online}
Figure~\ref{fig:streaming} reports the result for this task. We report results on the RotatE model with randomly initialized embeddings for new entities (RotatE) and the model with systematic initialization of new entity embeddings (RotatE+). We experiment with  $m = \{10\%, 30\%\}$ of previously seen edges and re-train on them. We find that not including previously seen edges leads to severe degradation of overall performance due to the model forgetting what it had learned in the past. We also report results with freezing the already seen entity representations and only learning representations for new entities (RotatE-Freeze). All models were trained till the validation set (containing both new and old triples) performance stopped improving. For our approach, we also report results for an oracle setting where we re-cluster all entities as new data arrives and re-estimate all parameters from scratch (instead of using \grinch and recomputing only required parameters (\S\ref{sub:open_world}). For both datasets, the offline-best results were obtained by RotatE (47.1 for \fb test-I, 48 for \wn). We report performance on the entire evaluation set (full) and also on the set containing the newly added edges (new).

The main summary of the results are (i) RotatE model converges to a much lower performance in the online setting losing at least 8 MRR points in \fb and at least 11 points in \wn. On \fb, we observe that the model prefers to learn new information more by sacrificing previously learned facts (2nd subfigure in figure \ref{fig:streaming}) (ii) In the freeze setting, the model performance deteriorates quickly after a certain point indicating saturation, i.e. it becomes hard for the model to learn new information about arriving entities by keeping the parameters of the existing entities fixed. (iii) On the full evaluation, RotatE+ performs better than RotatE showing that bad initialization deteriorates performance over time, however, there is still a large gap between the best performance (iv) Our approach almost matches our performance in oracle setting indicating the effectiveness of the online clustering and fast parameter approximation. (v) Lastly, we perform closest to the offline best results outperforming all variants of RotatE.
\section{Related Work}
% \vspace{-2mm}
\label{sec:related_work}
\noindent\textbf{Open-world KG completion}. \citet{shi2018open} consider the task of open-world KG completion. However, they use text descriptions to learn entity representations using convolutional neural networks. Our model does not use additional text data and we use very simple entity representations that helps us to perform well. \citet{tang2019learning} learns to update a KG with new links by reading news. Even though they handle adding or deleting new edges, they do not observe new entities. Lastly, none of them learn from similar entities using a CBR approach.

\noindent\textbf{Inductive representation learning on KGs}. Recent works \cite{teru2019inductive,wang2020entity} learn entity independent relation representations and hence allow them to handle unseen entities. However, they do not perform contextual reasoning by gathering reasoning paths from similar entities. Moreoever, in our open-world setting, we consider the more challenging setting, where new facts and entities are arriving in a streaming fashion and we give an efficient way of updating parameters using online hierarchical clustering. This allows our method to be applicable in settings where the initial KG is small and it grows continuously.

\noindent\textbf{Rule induction in knowledge graphs}. Classic work in inductive logic programming (ILP) \cite{muggleton1992inductive,quinlan1990learning} induce rules from grounded facts. However, they need explicit counter-examples which are not present in KBs and they do not scale to large KBs. Recent ILP approaches \cite{galarraga2013amie,galarraga2015fast} try to fix this deficiency by guessing counter examples from rules and making it more scalable. Statistical relational learning methods \cite{getoor2007introduction,kok2007,schoenmackers2010learning} and probabilistic logic approaches \cite{richardson2006markov,broecheler,wang2013programming} combine machine learning and logic to learn rules. However, none of these work derive reasoning rules dynamically from similar entities in the knowledge graph. 

\noindent\textbf{Bayesian non-parametric approaches for link-prediction}. There is a rich body of work in bayesian non-parametrics to automatically learn the latent dimension of entities \cite{kemp2006learning,xu2006infinite}. Our method does not learn latent dimension of entities, instead our work is non-parametric because it gathers reasoning paths from nearest neighbors and can seamlessly reason with new entities by efficiently updating parameters using online non-parametric hierarchical clustering.

\textbf{Embedding-based approach for link prediction}. We also compare to the more popular embeddings based models based on tensor factorization or neural approaches \cite{nickel2011three,bordes2013translating,dettmers2018convolutional,sun2019rotate}. Our simple approach which needs no iterative optimization outperforms most of them and performs comparably to the latest RotatE model. Moreover we outperform RotatE in the online experiments.

\textbf{CBR for KG completion}. There has been few attempts to apply CBR for knowledge management \cite{dubitzky1999viewing,bartlmae2000case},  however they do not do contextualized reasoning or consider online settings.
Our work is most closely related to the recent work of \citet{cbr}. However, since it does not take in to account the importance of each path, it suffers from low performance, with our model outperforming it in several benchmarks.
\section{Conclusion}
\vspace{-1.5mm}
\label{sec:conclusion}
We present a simple yet accurate approach for probabilistic case-based reasoning in knowledge bases. Our method is non-parametric, deriving reasoning rules dynamically from similar entities in the KB and is capable of handling new entities. We cluster similar entities together and estimate per-cluster parameters that measures the prior and precision of paths using simple count statistics. Our simple approach performs competitively to the best embeddings based models on several benchmarks and outperforms all models in the open-world setting.

\section*{Acknowledgements}
We thank anonymous reviewers and members of UMass IESL and NLP groups for helpful discussion and feedback. This work is funded in part by the Center for Data Science and the Center for Intelligent Information
Retrieval, and in part by the National Science Foundation under Grants No. IIS-1514053 and No. 1763618, and in part by the Chan Zuckerberg Initiative under the project Scientific Knowledge
Base Construction. Any opinions, findings and conclusions or recommendations expressed in this
material are those of the authors and do not necessarily reflect those of the sponsor.
\bibliographystyle{acl_natbib}
\bibliography{emnlp2020}
\newpage
\clearpage
\appendix

\section{Appendix}
\subsection{Entity Clusters}

Both clustering methods used in this paper, 
hierarchical agglomerative clustering (HAC) and
\textsc{Grinch} measure similarities between sets 
of clusters via a linkage function. In particular, 
we use average pairwise linkage. For two sets $A$ and $B$,
this is defined as:
\begin{align}
    \frac{1}{|A||B|} \sum_{a \in A} \sum_{a \in B} \textsf{sim}(a,b)
\end{align}

\subsection{Selecting Flat Clusterings}
\label{sub:appendix_flat_cluster}
A hierarchical clustering $T$ over the entities $\mathcal{V}$,
encodes a large number of flat partitions of the entities, 
often referred to as tree consistent partitions in the clustering 
literature. We select one of these tree consistent partitions 
using a threshold on the linkage function, $\tau$. The algorithm performs 
a breadth first search starting at the root node. The search stops
at any node for which the linkage is above the given value $\tau$. 
Pseudocode is given in Algorithm \ref{alg:flat clustering}.

\begin{algorithm}[t]
\caption{Select a flat clustering from a tree structure.}
\begin{algorithmic}[1]
\State \textbf{input:} {$\mathcal{V}$ : Entities , $root$: Root of tree, $\tau$: Threshold}
\State \textbf{output:} {$C_1,C_2,\dots,C_K$: A flat partition}
\State $frontier \gets [root]$
\State $result \gets \{\}$
\WHILE{$frontier$ is not empty}
\State $n \gets frontier.pop()$
\IF{$linkage(n) > \tau$}
\State $result \gets \{n\} \cup result$
\ELSE
\FOR{ $c$ in $n.children$}
\State $frontier.push(c)$
\ENDFOR
\ENDIF
\ENDWHILE
\State \textbf{return} $result$
\end{algorithmic}
\label{alg:flat clustering}
\end{algorithm}

\subsection{Number of Entity Updates Per Batch In Online Setting}

We analyze the number of entities that need to be re-clustered and
added in each round. We observe that it is significantly fewer 
than the number of entities in the KB. Note that an online method like
the one proposed in this paper just needs to run on the new and modified 
entities while a batch algorithm would need to run on the entire KB.

\begin{figure}
    \centering
    \includegraphics[width=0.48\textwidth]{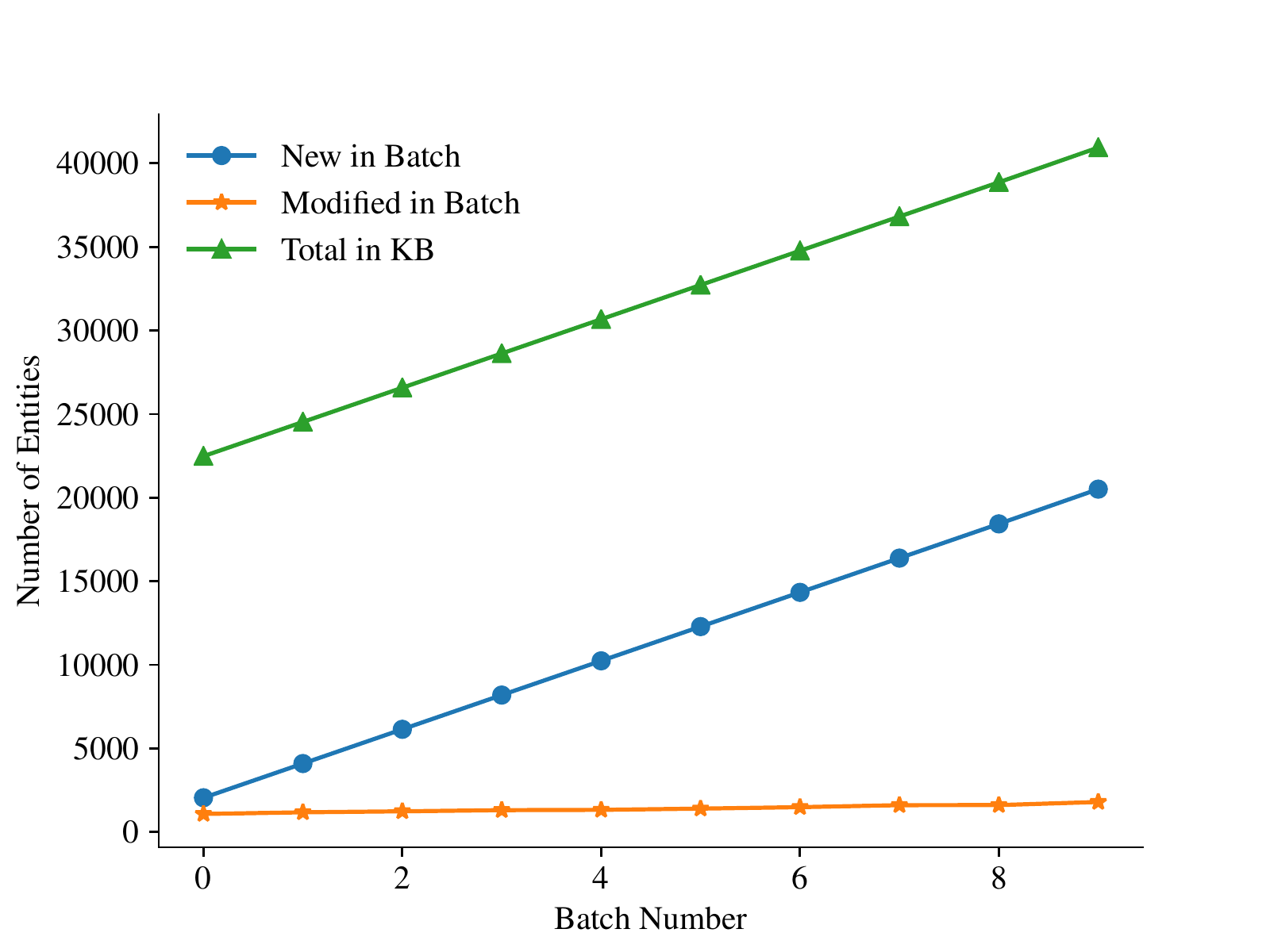}
    \caption{Number of entities added to KB in each batch and number of entities modified in each batch. These new and modified entities need to be updated in the clustering algorithm in each update.}
    \label{fig:streaming_up}
\end{figure}

\begin{table*}
\footnotesize
    \centering
    \begin{tabular}{c c c c c }
         \toprule
         &\bf People & \bf Professions & \bf Sports Org. & \bf Religious Entities  \\
         \midrule
         & Marvin Gay & Statistician & St. Louis Blues & Isalm \\
         At time &Shaquille O’Neal & Assoc. football manager  & Orlando Pirates  & Russian Orthodox church \\
         $t-1$ &Avril Lavinge & Structural Engineer & Sheffield Wednesday FC & Buddhism\\
         &Woody Harrleson & Financial backer & Malaya national football team & United Church of Christ \\
         \midrule
         \midrule
       At time & Elliot Smith & Harpsichordist &  Excelsior Rotterdam &  The Mormons  \\
        $t$ & Barbara Stanwick  & Child Actor &  Seattle Super Sonic &  Eastern Rite Catholic\\
         \bottomrule
    \end{tabular}
    \caption{Example Clusters discovered in online setting. We show the assignment of new entities to the clusters in the particular time step (below line).}
    \label{tab:appendix_cluster_eye_candy}
\end{table*}

\subsection{Finding entities for re-estimating parameters}
\label{sub:appendix_estimate_params}
\textit{Proposition}: Let $n$ denote the maximum length of a reasoning path considered by our model. For every new entity $e_i$ added to the KG, we need to recompute statistics for entities that lie within cycles of length up to $(n+1)$ starting from $e_i$.

We see from Eq \ref{eq:prior}, that the estimate for the prior for a \textit{path type} $p$ depends on $\mathcal{P}_{n}(e_c, r_q)$ i.e. the set of paths that lead from $e_c$ to entities that are connected to $e_c$ via relation $r_q$. WLOG, say $e_t$ is such an entity i.e. $(e_c, r_q, e_t) \in \mathcal{G}$. When a new entity/edge is added to the KG, this set of paths might increase. It is easy to see that the set $\mathcal{P}_{n}(e_c, r_q)$ is updated \textit{iff} a new path $p_{new}$ of length $\leq n$ appears between $e_c$ and $e_t$. In this case, the edges in $p_{new}$ would form a cycle with the edge $(e_c, r_q, e_t)$. The length of the cycle would be at most $len(p_{new}) + 1$ which in turn is at most of length $n+1$. This, to find entities for which the prior has changed after the addition of a new edge/entity, it is sufficient to find entities lying on cycles of length up to $n+1$ starting from the new entity/edge.

This mechanism for finding entities for re-computation is only approximate when computing the precision. We see from Eq \ref{eq:precision}, that the numerator depends on paths that lead to the answer entity (as with prior) while denominator depends on all $n$ length paths around $e_c$. So, if the numerator is ever to be increased, we would catch that update by the proposed cycle finding method. However, even if an entity does not lie on a cycle with the new edge/entity, if there is a path of length $n$ from $e_c$ to the new edge/entity, the denominator count would be incremented. Thus, the precision estimates for some entities might be an over-estimate of the path precision (had it been recomputed after new edges are added to the KB).

\begin{table}
\centering
\small
%\vspace{-2mm}
\begin{tabular}{ l  c c c}
\toprule
& \textbf{WN18RR} & \textbf{FB122} & \textbf{\nell}  \\ 
\midrule
  \textsc{hits}@1 & 0.422 & 0.694 & 0.296\\
  \textsc{hits}@3  & 0.461 & 0.739 & 0.405\\
  \textsc{hits}@10  & 0.508 & 0.779 & 0.502\\
  \textsc{MRR}   & 0.451 & 0.724 & 0.367\\
\bottomrule
\end{tabular}
\caption{Results on Validation set}
\label{tab:validation_results}
\end{table}

\begin{table}
\centering
\small
\begin{tabular}{ l  c c}
\toprule
& \textbf{WN18RR} & \textbf{NELL-995}   \\ 
\midrule
  \textsc{hits}@1 & 41.8 $\pm$ (5.7e-2) & 76.5 $\pm$ 2e-1\\
  \textsc{hits}@3  & 46.5 $\pm$ 0 & 85.2 $\pm$ 7e-2\\
  \textsc{hits}@10  & 51.3 $\pm$ (5.7e-2)& 89.5 $\pm$ 1.4e-2\\
  \textsc{MRR}   &  45 $\pm$ (5.7e-2) & 81.45 $\pm$ 2e-1\\
\bottomrule
\end{tabular}
\caption{Mean and Variance across different hyper-params}
\label{tab:diff_hyper_param_test}
\end{table}

\subsection{Example Clusters}
Table \ref{tab:appendix_cluster_eye_candy} shows some example of new entities arriving and getting assigned to their respective clusters by \grinch.

\subsection{Reproducibility Checklist}
\label{sub:repro_check}
\noindent\textbf{Computing Infrastructure}: All our experiments were run on a Xeon E5-2680 v4 @ 2.40GHz CPU with 128 GB RAM. No GPUs were needed for the experiments.   

The results on the validation set are reported in table \ref{tab:validation_results} and avg. of 3 runs are reported in table \ref{tab:diff_hyper_param_test}. The \nell does not come with a validation set, and therefore we selected 3000 edges randomly from the full NELL KB. As a result, many of the query relations were different from what was present in the splits of \nell and hence is not a good representative. However, we report test results for the best hyper-parameter values that we got on this validation set. 

The fixed number of parameters in our model are essentially the sparse non-learned entity vectors (which can be easily stored in COO format without taking much space). Other than that, our model is non-parametric with the number of parameters tied to the data.

For experiments on \textbf{\wn}:
\begin{itemize}
    \item Inference time: 18.9 queries/s (total of 6268 queries)
    \item Train time: around 20 mins.
    \item Best Hyper-parameters: 
    \begin{itemize}
        \item Number of nearest-neighbor entities ($K$): 40
        \item Number of paths from neighbors ($N$): 60
        \item Max length of path ($n$): 5
        \item Linkage for hierarchical clustering ($\lambda$): 0.25
    \end{itemize}
    \item Hyper-parameter method / bounds: Grid search
    \begin{itemize}
        \item $K$: [5, 10 , 15, 20, 30 , 40, 50]
        \item $N$: [5, 10, 20, 40, 60 , 80]
        \item $\lambda$: [0.25, 0.3, 0.35, 0.4, 0.45, 0.5, 0.6]
    \end{itemize}
     
\end{itemize}

For experiments on \textbf{\fb}:
\begin{itemize}
    \item Inference time: 
    \item Train time: around 90 mins
    \item Best Hyper-parameters: 
    \begin{itemize}
        \item Number of nearest-neighbor entities ($K$): 10
        \item Number of paths from neighbors ($N$): 80
        \item Max length of path ($n$): 3
        \item Linkage for hierarchical clustering ($\lambda$): 0.6
    \end{itemize}
    \item Hyper-parameter method / bounds: Grid search
    \begin{itemize}
        \item $K$: [5, 10 , 15, 20, 30 , 40, 50]
        \item $N$: [5, 10, 15, 25, 60, 80]
        \item $\lambda$: [0.4, 0.45, 0.5, 0.6, 0.65, 0.7, 0.75, 0.8, 0.95]
    \end{itemize}
\end{itemize}

For experiments on \textbf{\nell}:
\begin{itemize}
    \item Inference time: 9.05 queries/s (total of 2825 queries)
    \item Train time: around 90 mins
    \item Best Hyper-parameters: 
    \begin{itemize}
        \item Number of nearest-neighbor entities ($K$): 15
        \item Number of paths from neighbors ($N$): 25
        \item Max length of path ($n$): 3
        \item Linkage for hierarchical clustering ($\lambda$): 0.95
    \end{itemize}
    \item Hyper-parameter method / bounds: Random search
    \begin{itemize}
        \item $K$: [5, 10 , 15, 20, 30 , 40, 50]
        \item $N$: [5, 10, 20, 40, 60, 80]
        \item $\lambda$: [0.4, 0.45, 0.5, 0.6, 0.65, 0.7, 0.75]
    \end{itemize}
\end{itemize}

\end{document}